\documentclass[conference]{IEEEtran}

\usepackage{cite} 
\usepackage{amsmath,amssymb,amsfonts}
\usepackage{algorithmic}
\usepackage{graphicx}
\usepackage{textcomp}
\usepackage[dvipsnames,table,xcdraw]{xcolor}
\usepackage{tikz}
\usetikzlibrary{arrows.meta,positioning,fit,shapes.multipart,calc,backgrounds}
\usepackage{listings}
\usepackage{booktabs}
\usepackage{multirow}
\usepackage{float}

\definecolor{backcolour}{rgb}{0.95,0.95,0.92}
\definecolor{codegreen}{rgb}{0,0.6,0}

\begin{document}


\title{Toward a Modular Architecture for Embedded Agent Systems at the Edge}

\author{\IEEEauthorblockN{1\textsuperscript{st} Marcus Rüb}
\IEEEauthorblockA{\textit{Foresthub.Ai}\\
Villingen, Germany \\
m.rueb@foresthub.ai}

\and
\IEEEauthorblockN{2\textsuperscript{nd} Michael Gerhards}
\IEEEauthorblockA{\textit{Deloitte Consulting}\\
Düsseldorf, Germany \\
mgerhards@deloitte.de}
}
\maketitle 

\begin{abstract}
The rise of Large Language Models (LLMs) has enabled agentic AI capable of complex reasoning and tool use; however, deploying such autonomy in pervasive computing environments remains challenging due to the strict memory and energy constraints of embedded microcontrollers. Existing frameworks typically assume server-class resources or continuous connectivity, leaving a gap for deeply embedded systems. This paper proposes a modular reference architecture for \textit{Embedded Agent Systems} that bridges the divide between deterministic real-time control and agentic intelligence. 

We introduce a tiered design that decouples \textbf{On-Device Agents}---executing highly compressed neural networks and rule-based logic for low-latency, privacy-critical tasks---from \textbf{Cloud-Augmented Agents} that leverage Small Language Models (SLMs) for higher-level reasoning and planning. A key contribution is the integration of a cross-cutting \textbf{Governance Layer}, ensuring observability, policy enforcement, and safety across distributed fleets of autonomous devices. Rather than presenting purely empirical benchmarks, we analyze architectural design principles and trade-offs regarding latency, energy, and reliable execution in resource-constrained environments.
\end{abstract}


\section{Introduction}

Agentic AI systems, powered by Large Language Models (LLMs) and orchestration frameworks, have demonstrated remarkable capabilities in reasoning, planning, and automating complex workflows~\cite{xi2023rise, yao2022react}.  However, the deployment of such systems has largely been confined to cloud environments or high-performance data centers where memory, compute power, and energy are abundant. In sharp contrast, the physical world of pervasive computing is dominated by deeply embedded devices—microcontrollers (MCUs), sensor nodes, and low-power actuators—operating under strict resource constraints and often harsh connectivity conditions.

Currently, a disconnect exists between these two domains. Classical embedded development prioritizes deterministic control loops, static firmware, and hard real-time guarantees. Modern agentic AI, conversely, assumes dynamic context, abundant memory for history, and flexible tool use. Bridging this gap is non-trivial: while current hardware trends are promising, running a fully capable reasoning agent on a standard microcontroller remains technically infeasible due to the "memory wall"\cite{banbury2021mic, tinyml-microcontrollers}.

To address this, we argue for a differentiated approach based on hardware tiers. While powerful \textbf{Edge Devices} (e.g., gateways, MPUs) are increasingly capable of hosting quantized \textit{Small Language Models (SLMs \cite{SLM})} for local, autonomous reasoning, \textbf{Deeply Embedded MCUs} lack the resources for such models. Consequently, for the lowest tier of devices, the most viable path to "agentic" behavior is currently to act as an intelligent interface for \textbf{Cloud Agents}, or to execute highly optimized, non-generative TinyML models \cite{tinyml-microcontrollers, warden2019tflite}.

This paper proposes a modular architecture that unifies these heterogeneous deployment patterns. We present a framework where the definition of an "agent" adapts to the underlying hardware: ranging from a fully autonomous on-device agent on an edge gateway to a cloud-tethered sensor agent on a microcontroller. This flexibility allows system designers to balance latency, privacy, and reasoning depth dynamically.

Concretely, this paper makes the following contributions:
\begin{itemize}
    \item We analyze the hardware-specific constraints for embedded agents, distinguishing between SLM-capable edge devices and MCU-based cloud interfaces.
    \item We propose a \textbf{Modular Embedded Agent Architecture} that decouples the \textit{execution layer} (local vs. cloud) from the logical agent definition, allowing for hybrid deployment strategies.
    \item We introduce a cross-cutting \textbf{Governance Layer} as a mandatory component for managing distributed fleets of semi-autonomous agents, addressing observability and safety.
    \item We provide a conceptual evaluation of trade-offs across three application domains (Smart Agriculture, Predictive Maintenance, Smart Home), illustrating where local SLMs provide value versus where cloud offloading is superior.
\end{itemize}

The remainder of this paper is organized as follows. Section~II reviews the state of the art in SLMs and TinyML. Section~III details the unified architecture. Section~IV describes the functional core modules. Section~V analyzes the conceptual trade-offs, and Section~VI concludes the paper.

\section{State of the Art}

The landscape of AI on constrained devices is rapidly evolving. To contextualize the proposed architecture, it is necessary to distinguish between \textit{Edge AI} on microprocessor-based gateways (MPUs) and \textit{TinyML} on deeply embedded microcontrollers (MCUs).

\subsection{Small Language Models on Edge Gateways}
For edge devices possessing gigabytes of RAM (e.g., Raspberry Pi 5, NVIDIA Jetson, or industrial gateways), the execution of \textit{Small Language Models (SLMs)} has become feasible \cite{tinyml-microcontrollers, warden2019tflite}. Models such as TinyLlama (1.1B), Phi-3, or Qwen-1.5B ~\cite{zhang2024tinyllama, abdin2024phi3} are designed to balance reasoning capabilities with reduced parameter counts~\cite{huggingface-slm-overview}. 
Through 4-bit quantization and kernel optimization (e.g., via \textit{llama.cpp} or ONNX Runtime)\cite{tinyml-microcontrollers, warden2019tflite}, these models can run locally with acceptable token generation rates~\cite{dettmers2024qlora}. In our architecture, these devices serve as hosts for fully autonomous \textbf{On-Device Agents}, capable of local chain-of-thought reasoning without cloud connectivity.

\subsection{TinyML and Intelligence on Microcontrollers}
In contrast, deeply embedded MCUs (e.g., ESP32, STM32, Cortex-M) typically operate with strictly limited RAM (SRAM $< 512$\,KB) and Flash storage. Running billion-parameter models on such hardware is currently infeasible due to the "memory wall." 
Instead, intelligence in this tier relies on \textit{TinyML}: highly compressed, specialized neural networks for classification, anomaly detection, or keyword spotting using frameworks like TensorFlow Lite for Microcontrollers (TFLM)\cite{tinyml-microcontrollers, warden2019tflite}. 
While recent research explores "pico-scale" generative models (<10M parameters) for MCUs, their reasoning capacity remains limited. Therefore, for complex agentic tasks, MCUs are best utilized as intelligent interfaces that pre-process sensor data and delegate high-level reasoning to \textbf{Cloud Agents}.
Beyond pure inference, recent work has started to explore on-device training
and adaptation under these constraints. This includes continual and incremental
learning for TinyML using dataset distillation and model size adaptation
on microcontrollers~\cite{rueb2024incremental}, adaptive sparse
backpropagation schemes that reduce the cost of on-device training through
dynamic sparsity~\cite{rueb2024tinypropv2}, and
Grad-CAM-based streaming data prioritization that selectively retains only the
most informative samples for local training~\cite{rueb2024drip}. These
approaches are complementary to the architectural perspective of this paper:
while they optimize the learning algorithms themselves, our focus here is on
the system architecture required to host such learning-enabled agents.

\subsection{Embedded Agent Architectures}
Traditional multi-agent frameworks (e.g., JADE\cite{jade-book}) are too heavy-weight for embedded deployment. Recent efforts like \textit{micro-ROS} adapt robotic middleware for MCUs, leveraging real-time operating systems (RTOS) and DDS-based communication to enable node-like behavior on constrained hardware~\cite{bosch-microros}. 
However, micro-ROS focuses primarily on deterministic control and messaging, not on the LLM-driven "agentic" loops (plan-act-observe) seen in modern AI. Our work aims to fill this gap by defining an architecture that supports both deterministic control (via RTOS) and semantic reasoning (via SLM or Cloud).

\subsection{Tool Integration and Communication}
Modern agent systems rely on dynamic tool usage. While cloud agents utilize JSON-based APIs or the Model Context Protocol (MCP~\cite{mcp_protocol}), embedded devices typically communicate via lightweight telemetry protocols like MQTT or CoAP~\cite{mqtt-spec, coap-rfc}. 
A core challenge lies in bridging these worlds: mapping the verbose, text-heavy tool calls of an LLM/SLM to the binary or compact payload formats required by battery-powered MCUs. Existing solutions often rely on rigid gateways; our architecture proposes a more flexible \textit{Agent Interaction Interface} to standardize this translation.
\begin{figure*}[t] 
    \centering
    \includegraphics[width=\textwidth]{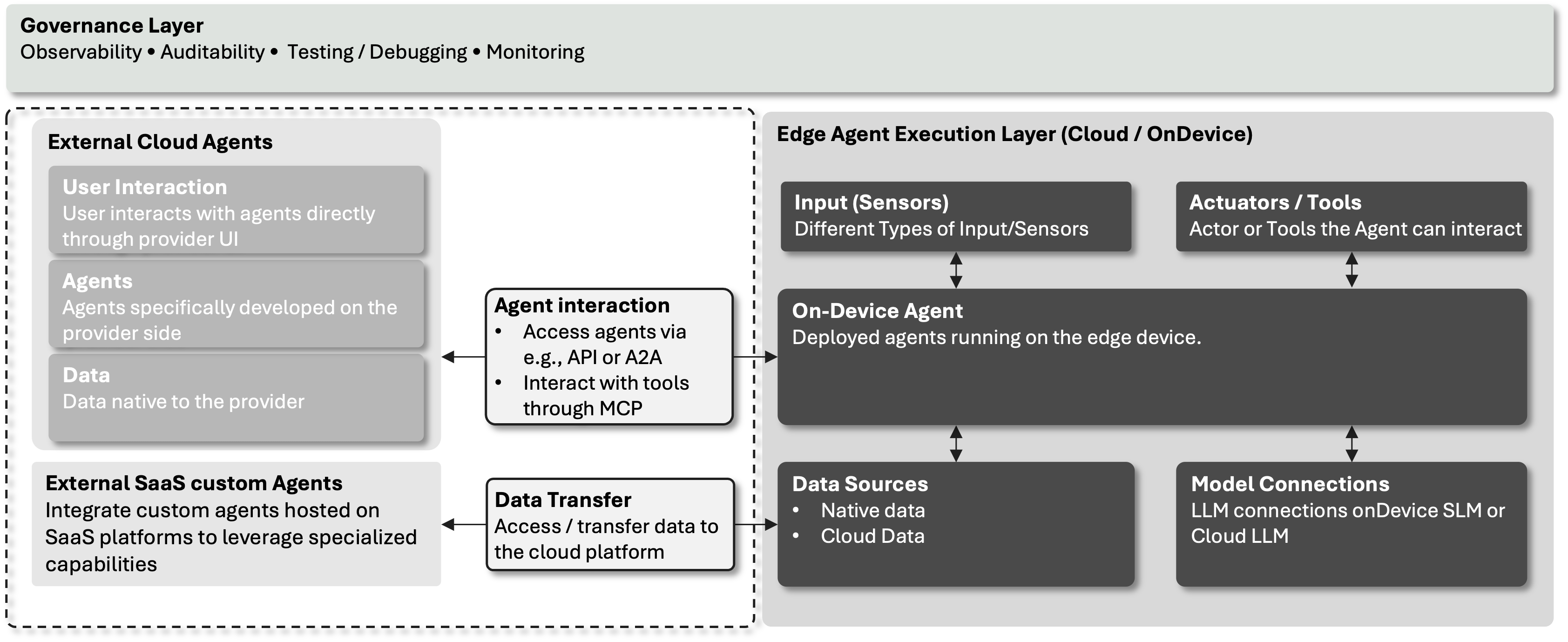} 
    
    \caption{Reference architecture showing the Edge Execution Environment. Depending on hardware class, the 'Agent Core' either runs a local SLM (Gateway) or delegates to External Cloud Agents (MCU).}
    \label{fig:arch-overview}
\end{figure*}
\section{Modular Embedded Agent Architecture}

The proposed architecture aims to bridge the gap between classical embedded development and modern agentic AI by providing a modular framework for resource-constrained devices. Its design follows the principle of \emph{composability}, where functional blocks can be combined depending on the deployment context and hardware limitations.

To address the hardware spectrum from microcontrollers to edge gateways, we distinguish two major deployment variants ("Flavors"): (i) the fully local \emph{Autonomous On-Device Agent}, and (ii) the \emph{Tethered Cloud Agent}, where the embedded system serves as an intelligent interface while intelligence is offloaded.

\subsection{Requirements for Embedded Agents}

Embedded agents in pervasive environments face a set of constraints and objectives that differ substantially from those of cloud-only agents. Derived from the literature on TinyML, IoT systems, and edge AI, and from practical experience with microcontroller-based deployments, we derive the following requirements:

\begin{itemize}
    \item \textbf{Low and predictable latency.}
    Many target applications (e.g., control loops, safety reactions, voice commands) require millisecond-scale response times and deterministic behaviour. The architecture must permit local perception--action loops without unpredictable delays introduced by networks or overloaded backends.

    \item \textbf{Energy efficiency.}
    Embedded agents often run on battery-powered or energy-harvesting devices. Both local computation and communication must be energy-aware, favouring lightweight models, duty-cycled sensing, and minimal uplink/downlink traffic.

    \item \textbf{Tight memory and compute budgets.}
    Microcontrollers typically offer only kilobytes of RAM and a few megabytes of Flash. The architecture must allow for compact model representations, static allocation strategies, and lean runtimes that fit within these limits while still providing useful agent behaviour.

    \item \textbf{Robustness to limited connectivity.}
    Connectivity in edge environments is often intermittent, lossy, or expensive. Embedded agents must remain functional when disconnected, with graceful degradation and local fallbacks, while exploiting the cloud opportunistically when links are available.

    \item \textbf{Privacy and security.}
    Sensor data may be sensitive (e.g., industrial telemetry, in-home signals). The architecture must support keeping data local whenever possible, enforce secure communication when data is offloaded, and provide mechanisms for authentication, authorisation, and integrity protection.

    \item \textbf{Interoperability and extensibility.}
    Edge deployments are heterogeneous: devices, vendors, and cloud providers vary. The architecture should separate transport, agent interaction, and tool access so that different protocols (e.g., MQTT, CoAP, HTTP) and tool ecosystems (e.g., MCP-based services) can be integrated without redesigning the agent core.

    \item \textbf{Observability and governance.}
    At scale, operators require visibility into fleets of embedded agents, including monitoring, logging, debugging, and policy enforcement. The architecture must accommodate a cross-cutting governance layer without embedding heavy management logic into each device.
\end{itemize}

\subsection{High-Level Reference Architecture}

Figure~\ref{fig:arch-overview} illustrates the proposed reference architecture. At its center lies the \emph{Edge Execution Environment}, which hosts the input processing, the agent core, and the output handling. 

This environment is designed to support two distinct configurations or "Flavors" based on the available hardware resources:

\begin{enumerate}
    \item \textbf{The Autonomous Gateway Agent (Flavor A):} Runs on capable edge hardware (MPUs). It executes SLMs locally for maximum privacy and autonomy.
    \item \textbf{The Tethered MCU Agent (Flavor B):} Runs on constrained microcontrollers. It relies on the cloud for reasoning but maintains local reflexes for safety.
\end{enumerate}

Regardless of the flavor, the architecture exposes an \emph{Agent Interaction} interface to allow collaboration via protocols like A2A or MCP, and a \emph{Data Transfer} interface for secure cloud synchronization. A cross-cutting \emph{Governance Layer} ensures auditability across all components.

\subsection{Flavor A: The Autonomous On-Device Agent}

This variant executes all essential components locally. It is designed for high-end embedded platforms (e.g., Edge Gateways, MPUs) to provide autonomy and privacy independent of cloud connectivity.

\textbf{Core Logic:} 
The agent integrates a quantized \textit{Small Language Model (SLM)} or a compressed neural network. These models are optimized via pruning and quantization to fit within the memory limits of the device (typically $>$512\,MB RAM). 
In addition to the SLM, the agent contains deterministic rule-based logic, a task manager to coordinate inputs and outputs, and a local tool layer providing access to embedded libraries (e.g., signal processing, filtering).

\textbf{Key Properties:}
\begin{itemize}
    \item \textbf{Low Latency:} Perception–action loops remain local, avoiding network round-trips.
    \item \textbf{Offline Capability:} Full functionality is maintained even without internet access.
    \item \textbf{Privacy:} Sensitive raw data never leaves the device.
\end{itemize}

\textbf{Trade-offs:} 
Model capacity is limited by onboard memory. These agents are well-suited for context-aware automation but may struggle with broad, open-domain knowledge compared to cloud models.

\subsection{Flavor B: The Tethered Cloud Agent (MCU)}

In this variant, the embedded device (e.g., ESP32, STM32) acts primarily as an intelligent interface. It captures inputs and forwards them to a cloud endpoint using lightweight protocols (MQTT, CoAP), reducing local computational load.

\textbf{Core Logic:} 
The on-device logic is minimized to signal conditioning, safety overrides, and communication handling. The complex reasoning is offloaded to a \textit{Coordinator Agent} in the cloud, which orchestrates specialized \textit{subagents} and Large Language Models (LLMs). This allows the use of powerful tools and massive knowledge bases that exceed the capacity of any microcontroller.

\textbf{Key Properties:}
\begin{itemize}
    \item \textbf{Scalability:} Intelligence can be upgraded centrally without re-flashing individual devices.
    \item \textbf{Advanced Reasoning:} Access to multi-billion parameter models and external SaaS tools via the coordinator agent.
    \item \textbf{Inter-Agent Communication:} Facilitates coordination between multiple distributed devices.
\end{itemize}

\textbf{Trade-offs:} 
The primary cost is latency (100--500\,ms round-trip) and dependence on connectivity. To mitigate risks, the local firmware typically includes a "safety fallback" mode that operates deterministically if the connection is lost. Furthermore, the transmission of sensor data requires mechanisms such as \textit{model armor} to protect both data integrity and the trustworthiness of responses.

\subsection{Shared Components and Governance}

Despite their differences, both variants share a common structural DNA (see Table~\ref{tab:shared-components}). The distinction lies in where these components are executed and how much computational complexity they can support.

\begin{table}[h]
\centering
\caption{Shared Components across Deployment Flavors}
\label{tab:shared-components}
\begin{tabular}{@{}lp{3.2cm}p{3.2cm}@{}}
\toprule
Component & Flavor A (On-Device) & Flavor B (Cloud/MCU) \\
\midrule
Input & Local sensors, GPIO & Forwarded via gateway \\
Agent Core & Local SLM Inference & Cloud LLM Coordinator \\
Tools & Local Libs, HAL & SaaS APIs, Databases \\
Interaction & Local Tool-Use & MCP / Cloud APIs \\
Output & Direct Actuation & Remote Commands \\
Security & Data stays local & Encryption \& Access Control \\
\bottomrule
\end{tabular}
\end{table}

In the \textbf{On-Device case}, inputs originate directly from local sensors and are processed by lightweight SLMs. Tools are limited to embedded libraries, and security focuses on device integrity.
In the \textbf{Cloud case}, inputs are forwarded to the infrastructure where powerful Coordinator Agents execute the reasoning. Tools include a wide range of APIs and external services. Security emphasizes access control and encryption during transmission.

\textbf{External Cloud Agents:} 
Both flavors can integrate with optional external agents hosted on third-party platforms. These are accessed via the agent-interaction interface and consume data exposed via the data-transfer interface, enabling an ecosystem of specialized capabilities.

\textbf{Governance Layer:} 
Finally, the architecture is bounded by a cross-cutting governance layer. This layer is orthogonal to the execution logic; it provides observability (logging), debugging support, and policy enforcement. In pervasive deployments, this ensures that fleets of embedded agents—whether local or cloud-tethered—can be operated within enterprise-level governance and compliance requirements.

\section{Core Modules of Embedded Agents}
\begin{figure}[t]
    \centering
    \includegraphics[width=0.48\textwidth]{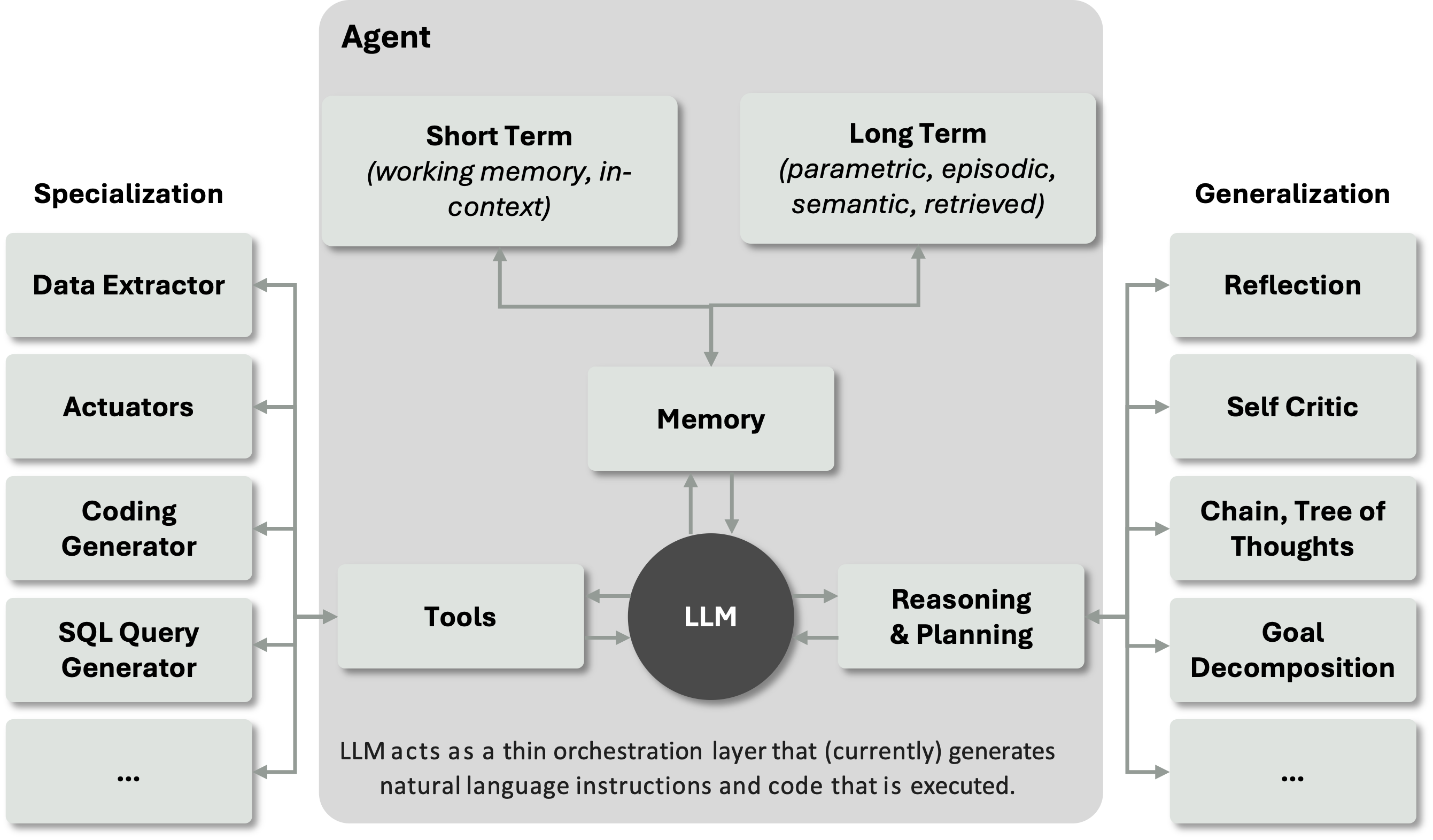}
    \caption{Internal structure of the Agent Core. The \textbf{LLM} acts as an orchestration layer connecting \textbf{Short/Long-Term Memory}, \textbf{Reasoning/Planning} modules (e.g., Chain of Thoughts), and \textbf{Tools} (e.g., Data Extractor, Actuators).}
    \label{fig:agent-core}
\end{figure}
Independent of whether an agent executes entirely on-device (Flavor A) or is cloud-tethered (Flavor B), its internal behaviour can be decomposed into a canonical set of functional modules. These modules form the \emph{Agent Core} within the Edge Execution Environment introduced in Section~III.

In this section, we describe the roles of these modules and discuss how their implementation differs depending on the available hardware resources, which are shown in Fig 2. 

\subsection{Perception}

The perception module is responsible for acquiring and pre-processing data from the environment. On embedded devices, this involves reading from GPIOs, analogue sensors, and digital interfaces such as I\textsuperscript{2}C, SPI, or UART.

\begin{itemize}
    \item \textbf{On MCUs (Flavor B):} Due to limited CPU, pre-processing is restricted to deterministic signal conditioning. Raw data is often compressed into statistical features (e.g., FFT magnitudes, RMS values) to minimize bandwidth before being sent to the cloud or a local TinyML classifier.
    \item \textbf{On Gateways (Flavor A):} The perception module can handle high-bandwidth modalities such as video streams or continuous audio buffers, performing local feature extraction to populate the context window of an SLM.
\end{itemize}

\subsection{Memory}

The memory module manages the agent's state. Unlike cloud agents that rely on massive vector databases, embedded agents face strict allocation limits. We distinguish three memory types:

\begin{enumerate}
    \item \textbf{Working Memory:} Stores current sensor readings and intermediate inference results. On MCUs, this is typically a static ring buffer in SRAM. On Gateways, this includes the \textit{KV-Cache} required for Transformer inference.
    \item \textbf{Episodic Memory:} Tracks recent events. MCUs may only store a "last state" flag, whereas Gateways can maintain a conversation history or a small local vector store (RAG-lite).
    \item \textbf{Configuration Memory:} Stores policies and parameters in Flash/NVM, ensuring persistence across reboots.
\end{enumerate}

\subsection{Planning}

The planning module decomposes high-level goals into executable actions. This is where the gap between the two flavors is most visible.

\begin{itemize}
    \item \textbf{Deterministic Planning (Flavor B):} On microcontrollers, planning is realized via lightweight, deterministic mechanisms such as Finite State Machines (FSMs) or Behavior Trees. These ensure predictable runtime behavior and satisfy hard real-time constraints.
    \item \textbf{Probabilistic Planning (Flavor A):} Gateways utilizing SLMs can perform dynamic task decomposition (e.g., "Chain-of-Thought" reasoning). Here, the planner generates a sequence of tool calls based on natural language instructions, allowing for flexible adaptation to unforeseen situations.
\end{itemize}

\subsection{Reasoning}

The reasoning module selects the next action based on perception and memory.

For \textbf{Flavor A (Gateway)}, this is the domain of quantized SLMs (e.g., Llama-3-8B-Quantized, Phi-3). These models run locally, mapping complex context descriptions to actions.

For \textbf{Flavor B (MCU)}, true generative reasoning is technically infeasible. Instead, we employ a \emph{Hybrid Reasoning} approach:
\begin{enumerate}
    \item \emph{Local Reflexes:} TinyML models (e.g., TFLite Micro) or rule engines handle immediate, safety-critical decisions (e.g., "Stop motor if vibration $>$ threshold").
    \item \emph{Remote Delegation:} For complex ambiguity, the agent serializes the state and requests a decision from the cloud coordinator.
\end{enumerate}

\subsection{Action}

The action module translates abstract decisions into physical effects. This includes driving actuators (motors, relays, LEDs) via Hardware Abstraction Layers (HAL) or invoking digital services.
Crucially, action execution must respect the real-time constraints of the embedded system. To avoid blocking the perception loop, actuator commands are typically offloaded to dedicated RTOS tasks, DMA channels, or interrupt service routines.

\subsection{Learning}

While fully-fledged training (Backpropagation) is computationally prohibitive on most embedded devices, the learning module enables lightweight adaptation.
\begin{itemize}
    \item \textbf{Parameter Tuning:} Agents can adjust sensitivity thresholds or running averages based on feedback.
    \item \textbf{Federated Learning:} Devices may compute gradient updates locally on a frozen model backbone and transmit only the updates to the server, preserving privacy ~\cite{kairouz2021federated}.
    \item \textbf{OTA Updates:} For major behavioral changes, the "learning" happens off-device, and the new policy is deployed via firmware-over-the-air (FOTA) updates.
\end{itemize}

\subsection{Tool Registry and Communication}

To interact with the world, the agent needs a catalogue of capabilities.
The \textbf{Tool Registry} maps semantic intents (e.g., "turn\_on\_light") to specific C-functions or driver calls. On constrained hardware, this registry is compiled statically to avoid dynamic dispatch overhead.

The \textbf{Communication Module} abstracts the transport layer. Whether using MQTT, CoAP, or raw serial, this module ensures that tool invocations and sensor data are serialized efficiently. In our architecture, this module acts as the adapter for the \textit{Model Context Protocol (MCP)}, allowing even simple MCUs to expose their sensors as "tools" to a larger AI ecosystem.

\subsection{Security and Safety}

Finally, the security module enforces boundaries.
\begin{itemize}
    \item \textbf{Safety Fallback:} A hard-coded, non-AI logic layer that validates every action before execution. It prevents the agent (whether local SLM or cloud LLM) from executing dangerous commands, such as overheating a boiler or moving a robotic arm outside safe limits.
    \item \textbf{Data Security:} On-device agents (Flavor A) benefit from inherent security as data stays local. Tethered agents (Flavor B) must employ TLS/DTLS encryption and strict device attestation to prevent man-in-the-middle attacks on the command stream.
\end{itemize}

\section{Evaluation (Conceptual Analysis)}

Since the proposed architecture addresses a spectrum of hardware capabilities—from constrained microcontrollers to powerful edge gateways—a direct empirical comparison is complex. Instead, we focus on a conceptual evaluation of the design trade-offs inherent to the two deployment flavors introduced in Section~III: the \textbf{Autonomous Gateway Agent (Flavor A)} and the \textbf{Tethered MCU Agent (Flavor B)}.

We analyze these variants across (i) theoretical resource constraints, (ii) representative application scenarios, and (iii) a summary of strategic trade-offs.

\subsection{Resource Considerations}

\textbf{Latency and Determinism.}
For real-time control, latency is the critical metric.
\begin{itemize}
    \item \textit{Flavor A (Gateway):} Local SLM inference, while private, is computationally heavy. Generating a reasoning response can take 200\,ms to several seconds depending on hardware (e.g., NPU vs. CPU). However, the \textit{reflex layer} (perception-action loop) remains fast ($<10$\,ms).
    \item \textit{Flavor B (MCU):} Local reflexes via TinyML are extremely fast ($<1$\,ms). However, any complex reasoning requires a network round-trip, adding unpredictable jitter (100--1000\,ms) and making this flavor unsuitable for high-frequency cognitive loops.
\end{itemize}

\textbf{Energy Consumption.}
The energy profile differs fundamentally between computation and communication.
\begin{itemize}
    \item \textit{Computation Dominant (Flavor A):} Running a 3-billion parameter model locally consumes significant power (Watts), limiting battery life.
    \item \textit{Communication Dominant (Flavor B):} The MCU consumes milliwatts during sleep/sensing but experiences power spikes during Wi-Fi/LTE transmission. For sporadic reasoning tasks (e.g., once per hour), Flavor B is far more energy-efficient. For continuous dialogue or analysis, local processing (Flavor A) may become more efficient by avoiding constant radio uptime.
\end{itemize}

\textbf{Memory and The "Memory Wall".}
Memory is the primary constraint dictating the choice of flavor.
\begin{itemize}
    \item \textit{Flavor B:} With RAM in the kilobyte range, MCUs cannot host generative models. They are restricted to static buffers and TinyML classifiers.
    \item \textit{Flavor A:} Requires gigabytes of RAM to hold model weights and the KV-cache for context. This necessitates higher BOM (Bill of Materials) costs, making it less viable for massive sensor deployments.
\end{itemize}

\textbf{Connectivity and Privacy.}
\begin{itemize}
    \item \textit{Flavor A} offers superior privacy and robustness. Since the SLM runs locally, raw video or audio data never leaves the premise, and the agent functions fully in air-gapped environments.
    \item \textit{Flavor B} relies on the cloud. While data can be encrypted, metadata and usage patterns are visible to the provider. Furthermore, loss of connectivity degrades the agent to a "lobotomized" state where only basic safety reflexes function.
\end{itemize}

\subsection{Use-Case Scenarios}

To illustrate the selection of the appropriate flavor, we discuss three representative scenarios.

\paragraph{Smart Agriculture (Flavor B: Tethered).}
Consider a soil monitoring system across a large field. Deploying expensive gateways at every measurement point is cost-prohibitive. Instead, low-cost MCUs (Flavor B) measure moisture and control valves based on simple local rules (reflex). For strategic decisions (e.g., "Conserve water due to upcoming drought"), they periodically fetch instructions from a Cloud Coordinator.
\textit{Verdict: Cost and energy constraints favor Flavor B.}

\paragraph{Predictive Maintenance (Hybrid).}
In an industrial setting, vibration sensors monitor high-value machinery. An MCU (Flavor B) performs continuous high-frequency sampling and FFT analysis locally. Only when a statistical anomaly is detected, a snapshot of the data is sent to an on-premise Edge Gateway (Flavor A) or Cloud.
\textit{Verdict: Hybrid approach; MCU for sensing, higher tier for diagnostics.}

\paragraph{Privacy-First Smart Home (Flavor A: Autonomous).}
A voice assistant for controlling home infrastructure handles sensitive data. Using a Tethered Agent (Flavor B) would stream audio to the cloud, raising privacy concerns. An Autonomous Gateway Agent (Flavor A) running a local speech-to-text and SLM pipeline ensures that voice data remains within the home network, functioning even during internet outages.
\textit{Verdict: Privacy and reliability requirements favor Flavor A.}

\subsection{Trade-Off Summary}

Table~\ref{tab:tradeoffs} summarizes the architectural decisions. It highlights that there is no single "best" embedded agent; rather, the choice depends on the specific constraints of the deployment environment.

\begin{table}[h]
\centering
\caption{Strategic Trade-offs: Flavor A (Gateway) vs. Flavor B (MCU)}
\label{tab:tradeoffs}
\begin{tabular}{@{}lcc@{}}
\toprule
Metric & \textbf{Flavor A (Gateway)} & \textbf{Flavor B (MCU)} \\
\midrule
\textbf{Hardware Cost} & High (\$50 -- \$200+) & Low (\$2 -- \$10) \\
\textbf{Reasoning} & Local Generative AI & Cloud LLM / Local Reflex \\
\textbf{Latency} & High Inference Time & Network Latency Jitter \\
\textbf{Privacy} & Air-Gapped / High & Cloud-Dependent \\
\textbf{Energy} & High Compute Load & High Radio Load \\
\textbf{Resilience} & Fully Autonomous & Vulnerable to Disconnects \\
\bottomrule
\end{tabular}
\end{table}

Overall, the analysis suggests that \textbf{Flavor A} is essential for tasks requiring deep reasoning, privacy, or offline reliability, whereas \textbf{Flavor B} allows "agentic" capabilities to scale to billions of low-cost endpoints by leveraging the cloud. The proposed architecture supports both modes within a unified governance framework.

\section{Discussion}

The proposed architecture provides a unified blueprint for bringing agentic behavior to the edge. By distinguishing between autonomous gateways (Flavor A) and tethered microcontrollers (Flavor B), we address the heterogeneity of the physical world. In this section, we reflect on the implications of this design, its limitations, and the emerging trends that will shape its evolution.

\subsection{Benefits and Design Implications}

\textbf{Unified Abstraction over Heterogeneous Hardware.}
The primary contribution of this work is the decoupling of the "Agent Concept" from the underlying execution hardware. By formally modeling the \textit{Edge Execution Environment}, we allow a system architect to treat a \$2 microcontroller and a \$100 edge gateway as nodes in the same agentic system. This supports principled design decisions: latency-critical reflexes are assigned to Flavor B (MCU), while privacy-sensitive reasoning is assigned to Flavor A (Gateway), without breaking the overall system topology.

\textbf{Modularity as an Enabler.}
The decomposition into core modules (Perception, Memory, Planning, etc.) provides a technology-agnostic template.
For instance, the \textit{Planning Module} can be instantiated as a deterministic Finite State Machine on an MCU or as a probabilistic Chain-of-Thought engine on a Gateway. This modularity prevents the "feature creep" often seen in IoT, where lightweight devices are overloaded with complex software stacks they cannot sustain.

\textbf{Governance as a First-Class Citizen.}
By introducing a cross-cutting Governance Layer, the architecture addresses the "Black Box" problem of AI at the edge. Whether an action is triggered by a local neural network or a cloud command, the governance layer enforces a uniform audit trail. This is a prerequisite for deploying autonomous agents in regulated industries like energy or healthcare.

\subsection{Current Limitations and Open Challenges}

\textbf{The Hardware-Cost Trade-off.}
While Flavor A (Autonomous Gateway) offers the ideal "privacy + intelligence" package, it remains cost-prohibitive for ubiquitous sensing. Even compressed SLMs require hardware (NPU, $>$1GB RAM) that exceeds the BOM cost of typical IoT endpoints. Consequently, Flavor B (Tethered) will remain the dominant paradigm for mass deployment in the near term, necessitating robust cloud connectivity.

\textbf{The Semantic Gap in Communication.}
A significant technical challenge lies in the \textit{Agent Interaction} interface. Cloud agents and SLMs typically communicate via verbose, text-based protocols (e.g., JSON schemas in MCP). Mapping these rich semantic structures to the binary, bandwidth-constrained protocols of microcontrollers (e.g., CBOR over CoAP) requires efficient translation layers that are not yet standardized.

\textbf{Lack of Empirical Benchmarks.}
As this is a conceptual architectural proposal, we have not provided end-to-end performance metrics. Validating the "Safety Fallback" mechanisms and measuring the exact energy overhead of the proposed "Hybrid Reasoning" approach across different hardware platforms remains a critical task for future work.

\subsection{Lessons Learned and Emerging Trends}

\textbf{From "Cloud-First" to "Tiered" Intelligence.}
The most vital lesson is that pushing full agentic AI to every node is neither feasible nor necessary. A tiered approach—combining billions of "reflexive" sensors (Flavor B) with millions of "reasoning" gateways (Flavor A)—offers a pragmatic path forward.

\textbf{Emergence of NPU-equipped MCUs.}
We observe a trend towards "TinyML-native" silicon (e.g., Arm Cortex-M55/M85). As these chips become commodities, the boundary between Flavor A and B will blur. Future MCUs may run "Nano-SLMs" capable of limited linguistic reasoning, allowing Flavor B agents to gain partial autonomy and reducing reliance on the cloud.

\textbf{Standardization of Tool Use.}
The rapid adoption of standards like the Model Context Protocol (MCP) suggests that "Tool Use" will become the universal interface for embedded systems. In the future, firmware developers may no longer write manual API documentation, but instead expose their sensors as MCP-compatible tools, making them instantly discoverable by both local and cloud agents.

\section{Conclusion and Future Work}

This paper has proposed a modular reference architecture for \textit{Embedded Agent Systems}, explicitly tailored to the heterogeneous reality of the edge. Motivated by the disconnect between resource-heavy Agentic AI and resource-constrained embedded development, we articulated a set of requirements centered on latency, energy, privacy, and governance.

A core contribution of this work is the tiered architectural model that unifies two distinct deployment paradigms: the \textbf{Autonomous Gateway Agent (Flavor A)}, which leverages local SLMs for privacy-critical and offline reasoning, and the \textbf{Tethered MCU Agent (Flavor B)}, which acts as a cost-effective, deterministic interface to cloud intelligence. By decomposing the agent into functional modules—from perception to planning—and wrapping them in a cross-cutting \textbf{Governance Layer}, we provided a blueprint that allows system architects to balance the trade-offs between local autonomy and cloud-scale reasoning within a single fleet.

Our conceptual evaluation highlights that there is no "one-size-fits-all" solution. While on-device agents offer superior latency and privacy, the "memory wall" currently restricts generative reasoning to higher-end gateways. Consequently, the proposed architecture emphasizes \textit{hybrid composability}, enabling simple microcontrollers to participate in agentic workflows via standardized interfaces and robust safety fallbacks.

Future work will proceed along three strategic directions:
\begin{enumerate}
    \item \textbf{Reference Implementation:} We plan to develop an open-source reference stack validating both flavors: a Python/C++ implementation for SLM-based gateways (e.g., Raspberry Pi with NPU) and a C-based micro-ROS/FreeRTOS implementation for tethered MCUs (e.g., ESP32-S3), measuring end-to-end latency and energy overheads.
    \item \textbf{Bridging the Semantic Gap:} We aim to investigate efficient protocol translation layers that map the verbose, text-based tool definitions of modern AI (e.g., MCP JSON schemas) to binary, bandwidth-efficient IoT protocols (e.g., CBOR/CoAP) without losing semantic precision.
    \item \textbf{Governance Tooling:} We intend to prototype the proposed governance layer, focusing on distributed tracing and "model armor" mechanisms that can detect and block hallucinated commands before they trigger physical actuators.
\end{enumerate}

Taken together, these steps aim to transition embedded agents from a theoretical concept toward practical, industry-ready deployments, paving the way for a ubiquitous computing landscape where every device—from the smallest sensor to the smartest gateway—can act as an intelligent agent.

\bibliographystyle{IEEEtran}
\bibliography{references} 

@article{huggingface-slm-overview,
  author       = {Jokah, HuggingFace},
  title        = {Small Language Models: An Overview},
  journal      = {HuggingFace Blog},
  year         = {2025},
  url          = {https://huggingface.co/blog/small-language-models},
  note         = {Accessed: 2025-09-28}
}

@article{xi2023rise,
  title={The rise and potential of large language model based agents: A survey},
  author={Xi, Zhiheng and Chen, Wenxiang and Guo, Xin and He, Yi and Ding, Yi and Hong, Boyang and Zhang, Shihan and Wang, Junzhe and Jin, Huhan and Zhou, Enyu and others},
  journal={arXiv preprint arXiv:2309.07864},
  year={2023},
  note={Comprehensive survey on LLM Agents}
}

@inproceedings{yao2022react,
  title={ReAct: Synergizing Reasoning and Acting in Language Models},
  author={Yao, Shunyu and Zhao, Jeffrey and Yu, Dian and Du, Nan and Shafran, Izhak and Narasimhan, Karthik and Cao, Yuan},
  booktitle={International Conference on Learning Representations (ICLR)},
  year={2023}
}

@inproceedings{banbury2021mic,
  title={Micronets: Neural network architectures for deploying tinyml applications on commodity microcontrollers},
  author={Banbury, Colby and Zhou, Chuteng and Fedorov, Igor and Matas, Ramon and Thakker, Urmish and Gope, Dibakar and Janapa Reddi, Vijay and Mattina, Matthew and Whatmough, Paul},
  booktitle={Proceedings of the Machine Learning and Systems (MLSys)},
  volume={3},
  pages={517--532},
  year={2021}
}

@article{zhang2024tinyllama,
  title={TinyLlama: An Open-Source Small Language Model},
  author={Zhang, Peiyuan and Zeng, Guangtao and Wang, Tianduo and Lu, Wei},
  journal={arXiv preprint arXiv:2401.02385},
  year={2024}
}

@article{abdin2024phi3,
  title={Phi-3 Technical Report: A Highly Capable Language Model Locally on Your Phone},
  author={Abdin, Marah and others},
  journal={arXiv preprint arXiv:2404.14219},
  year={2024},
  note={Microsoft Technical Report}
}

@inproceedings{dettmers2024qlora,
  title={QLoRA: Efficient Finetuning of Quantized LLMs},
  author={Dettmers, Tim and Pagnoni, Artidoro and Holtzman, Ari and Zettlemoyer, Luke},
  booktitle={Advances in Neural Information Processing Systems (NeurIPS)},
  volume={36},
  year={2024}
}

@book{warden2019tflite,
  title        = {TinyML: Machine Learning with TensorFlow Lite on Microcontrollers},
  author       = {Warden, Pete and Situnayake, Daniel},
  publisher    = {O’Reilly Media},
  year         = {2019}
}

@inproceedings{tinyml-microcontrollers,
  author       = {David, Ronny and others},
  title        = {TensorFlow Lite Micro: Embedded Machine Learning for TinyML Systems},
  booktitle    = {Proceedings of MLSys},
  year         = {2021}
}

@inproceedings{bosch-microros,
  title={Micro-ROS: The move of ROS 2 to microcontrollers},
  author={Lange, Ralph and others},
  booktitle={IEEE International Conference on Autonomous Robot Systems and Competitions (ICARSC)},
  year={2020}
}

@techreport{mqtt-spec,
  title={MQTT Version 5.0},
  author={Banks, Andrew and Gupta, Rahul},
  institution={OASIS Standard},
  year={2019}
}

@techreport{coap-rfc,
  title        = {The Constrained Application Protocol ({CoAP})},
  author       = {Shelby, Zach and Hartke, Klaus and Bormann, Carsten},
  institution  = {Internet Engineering Task Force},
  number       = {RFC 7252},
  year         = {2014}
}

@book{jade-book,
  author    = {Bellifemine, Fabio and Caire, Giovanni and Greenwood, Dominic},
  title     = {Developing Multi-Agent Systems with JADE},
  publisher = {John Wiley \& Sons},
  year      = {2007},
  isbn      = {978-0-470-05747-6}
}

@inproceedings{rueb2024incremental,
  author    = {R{\"u}b, Marcus and Tuchel, Philipp and Sikora, Axel and Mueller-Gritschneder, Daniel},
  title     = {A Continual and Incremental Learning Approach for {TinyML} On-device Training Using Dataset Distillation and Model Size Adaption},
  booktitle = {Proceedings of the IEEE 7th International Conference on Industrial Cyber-Physical Systems (ICPS)},
  year      = {2024},
  pages     = {1--8},
  doi       = {10.1109/ICPS59941.2024.10639989},
  organization = {IEEE}
}

@inproceedings{rueb2024tinypropv2,
  author    = {R{\"u}b, Marcus and Sikora, Axel and Mueller-Gritschneder, Daniel},
  title     = {Advancing On-Device Neural Network Training with {TinyPropv2}: Dynamic, Sparse, and Efficient Backpropagation},
  booktitle = {Proceedings of the International Joint Conference on Neural Networks (IJCNN)},
  year      = {2024},
  pages     = {1--8},
  doi       = {10.1109/IJCNN60899.2024.10650122},
  organization = {IEEE}
}

@inproceedings{rueb2024drip,
  author    = {R{\"u}b, Marcus and Konegen, Daniel and Sikora, Axel and Mueller-Gritschneder, Daniel},
  title     = {{DRIP}: {DR}op {unImportant} Data Points -- Enhancing Machine Learning Efficiency with {Grad-CAM}-Based Streaming Data Prioritization for On-Device Training},
  booktitle = {Proceedings of the International Joint Conference on Neural Networks (IJCNN)},
  year      = {2025},
  doi       = {10.1109/IJCNN64981.2025.11228149},
  organization = {IEEE}
}

@article{kairouz2021federated,
  title        = {Advances and Open Problems in Federated Learning},
  author       = {Kairouz, Peter and McMahan, H. Brendan and Avent, Brendan and Bellet, Aur{\'e}lien and others},
  journal      = {Foundations and Trends in Machine Learning},
  volume       = {14},
  number       = {1--2},
  pages        = {1--210},
  year         = {2021}
}

@misc{SLM,
  title        = {Small Language Models are the Future of Agentic AI},
  author       = {Belcak, Peter and Heinrich, Greg and Diao, Shizhe and Fu, Yonggan and Dong, Xin and Muralidharan, Saurav and Lin, Yingyan Celine and Molchanov, Pavlo},
  year         = {2025},
  eprint       = {2506.02153},
  archivePrefix= {arXiv},
  primaryClass = {cs.AI},
  url          = {https://arxiv.org/abs/2506.02153},
  doi          = {10.48550/arXiv.2506.02153}
}

@misc{mcp_protocol,
  title={Model Context Protocol (MCP): An open standard for connecting AI models to data and tools},
  author={Anthropic and Open Source Contributors},
  year={2024},
  howpublished={\url{https://modelcontextprotocol.io}}
}

\end{document}